\documentclass{article}

\usepackage{arxiv}

\usepackage[utf8]{inputenc} 
\usepackage[T1]{fontenc}    
\usepackage{hyperref}       
\usepackage{url}            
\usepackage{booktabs}       
\usepackage{amsfonts}       
\usepackage{nicefrac}       
\usepackage{microtype}      
\usepackage{lipsum}
\usepackage{graphicx}
\usepackage{csquotes}
\usepackage[printwatermark]{xwatermark}
\usepackage{xcolor}
\usepackage{epigraph}
\usepackage{wrapfig}
\usepackage{xparse}

\def\signed #1{{\leavevmode\unskip\nobreak\hfil\penalty50\hskip2em
  \hbox{}\nobreak\hfil(#1)%
  \parfillskip=0pt \finalhyphendemerits=0 \endgraf}}

\newsavebox\mybox
\newenvironment{aquote}[1]
  {\savebox\mybox{#1}\begin{quote}}
  {\signed{\usebox\mybox}\end{quote}}

\newcommand{\theDate}{February 20, 2020}
\newcommand{\topHeader}{}
\graphicspath{ {./images/} }
\SetBlockThreshold{0}

\makeatletter
\renewcommand*{\@fnsymbol}[1]{}
\makeatother
\title{A Road Map to Strong Intelligence}

\author{Philip Paquette \\
        pcpaquette@gmail.com \\
        Toronto, Canada}

\begin{document}
\maketitle

\begin{abstract}
I wrote this paper because technology can really improve people's lives. With it, we can live longer in a healthy body, save time through increased efficiency and automation, and make better decisions. To get to the next level, we need to start looking at intelligence from a much broader perspective, and promote international interdisciplinary collaborations. \
Section~\ref{sec:intelligence_is_social} of this paper delves into sociology and social psychology to explain that the mechanisms underlying intelligence are inherently social. \
Section~\ref{sec:levels_of_intelligence} proposes a method to classify intelligence, and describes the differences between weak and strong intelligence. \
Section~\ref{sec:chinese_room} examines the Chinese Room argument from a different perspective. It demonstrates that a Turing-complete machine cannot have strong intelligence, and considers the modifications necessary for a computer to be intelligent and have understanding. \
Section~\ref{sec:singleton} argues that the existential risk caused by the technological explosion of a single agent should not be of serious concern. \
Section~\ref{sec:control_problem} looks at the AI control problem and argues that it is impossible to build a super-intelligent machine that will do what it creators want. By using insights from biology, it also proposes a solution to the control problem. \
Section~\ref{sec:discussion} discusses some of the implications of strong intelligence. \
Section~\ref{sec:challenges_deep_learning} lists the main challenges with deep learning, and asserts that radical changes will be required to reach strong intelligence. \
Section~\ref{sec:brain_is_prediction_machine} examines a neuroscience framework that could help explain how a cortical column works. \
Section~\ref{sec:roadmap} lays out the broad strokes of a road map towards strong intelligence. \
Finally, section~\ref{sec:the_future} analyzes the impacts and the challenges of greater intelligence.
\end{abstract}

\section{Intelligence is Social}
\label{sec:intelligence_is_social}

\begin{aquote}{P. Paquette}
\textit{Intelligence is inherently social. Social interactions are not the results of intelligent behaviour, but the essence from which intelligence emerges.}
\end{aquote}

For clarity, I am broadly defining intelligence as \enquote{the ability to adapt to change}; an agent as \enquote{an entity that can perform interactions with other agents}; and an interaction as \enquote{any particular way an agent can be affected by another agent or by reality}. An agent, which can be composed of other agents, is anything, small or large, that can interact with other agents. A cell, a neuron, and a human are all examples of agents.

An interaction happens when an agent is influenced by another agent, or by reality. Interactions can be classified as (1) sensorial, (2) social, (3) cognitive, and (4) emotional. A \textbf{sensorial interaction} (a.k.a. qualia) happens when an agent is influenced through its senses. A \textbf{social interaction} happens when an agent is influenced by another agent. A \textbf{cognitive interaction} happens when an agent is influenced through its cognition (e.g. reading). A \textbf{emotional interaction} happens when an agent is influenced by its internal state of mind. These classifications are not exclusive. For instance, touching a table could be a sensorial and cognitive interaction, while discussing it could be a social, cognitive, and emotional interaction.

At its core, intelligence uses two key mechanisms, namely the acquisition of knowledge and the reduction of uncertainty, which both depend on social interactions.


\subsection{Acquisition of Knowledge}

Knowledge can either be acquired (\textit{knowledge acquisition}) or discovered (\textit{knowledge discovery} or \textit{creativity}). Knowledge acquisition can be performed through any type of interactions, but social interactions are the key components needed to bootstrap the minimal understanding of the world required to acquire additional knowledge.

\textbf{Data Is Not Knowledge.} It is crucial to understand the counterintuitive concept that data, by itself, is not sufficient to acquire knowledge. For instance, a book in Russian or Chinese might contain a lot of knowledge, but unless you are able to link that knowledge to the concepts you already know, you will not be able to acquire any knowledge. Language is a social construction and words have no intrinsic meaning \cite{cooley1902human, mead1934mind, clair1982language}. Knowledge can only be acquired through language if someone, who already knows what the agreed meaning of words is, tells you. The initial acquisition of knowledge can only be performed through social interactions. In other words, you cannot figure out the meaning given to words by everyone else, unless you have an interaction with them. This has profound implications and implies that a system cannot acquire any knowledge by simply crawling through a lot of data. 

This concept goes much further. Objects also have no intrinsic meaning but the meaning we give them, and the interpretation of reality is a social construction \cite{berger1967social, delamater2018social}. For instance, what is a chair if you have never seen a human and do not understand what legs and a torso are? What is a fork if you don't understand that humans have arms, fingers, a mouth and need to eat? A child can only acquire the initial knowledge required to interpret the world through social interactions. It is simply impossible for any system to acquire an interpretation of the world that is compatible with ours, if that system has no interactions with humans.

\textbf{Transmission Speed.} If an alien civilization wanted to transfer us all their knowledge, how fast would the acquisition of knowledge happen? The transfer cannot happen instantly, and the acquisition speed would be limited by the slowest of their transmission speed and our processing speed. In other words, the transfer is limited by the slowest of their writing speed and our reading speed, or their speaking speed and our listening speed.

It is worth noting that the speed of transfer would gradually increase. The more things one knows, the easier it is to link additional concepts to one's existing knowledge. Moreover, without social interactions (i.e. being taught), the knowledge could only be discovered, and the transfer would happen much slower, if at all.

Assuming we can emulate brains in computers, could knowledge be transferred instantly? Yes, in theory, we could transfer knowledge almost instantly, but arguably there is no new knowledge being acquired. The acquisition of new knowledge between two emulated brains would still require social interactions (i.e. synchronization), and the transfer would be limited by either the transmission speed or the processing speed.

\textbf{Unshared Knowledge.} If language and the interpretation of reality are social constructions, knowledge has to be shared to be meaningful. Is a word really a word if you are the only one that knows it? Can someone invent new theories in their head, or would others think that person is completely disconnected from reality? Moreover, if two agents are only having interactions with each other, they will slowly synchronize their knowledge, and it would be hard for an agent to become significantly better than the other without having external interactions.

\textbf{Optimization Perspective.} From an optimization perspective, the acquisition or discovery of knowledge does not reduce the uncertainty of a system, but constantly reshapes the optimization landscape.


\subsection{Reduction of Uncertainty}

In a world with uncertainty, it is a necessity to have a mechanism that can find invariant structures, contrast possible events to determine how probable they are, and make probabilistic predictions about what is going to happen. Agents are able to reduce the uncertainty and update their beliefs about the world using social interactions, namely competition, collaboration, and conflict. These constant interactions are required to function in a world with limited resources.

For instance, humans can argue and work together to determine a probable outcome, and then update their beliefs depending on what happened. Similarly, neurons in a neural network are in constant cooperation and competition, and then update their beliefs about the world to learn. A similar logic can be applied to the mind, where conflicting thoughts are simultaneously present, to determine what the mind is thinking.

\textbf{Optimization Perspective.} From an optimization perspective, the social interactions between the agents reduces the uncertainty of the system and updates their belief of the world, based on the constantly changing optimization landscape caused by knowledge acquisition.

The concept that intelligence is social is a generalization of Minsky's Society of Mind \cite{minsky1988society}.


\section{Levels of Intelligence}
\label{sec:levels_of_intelligence}

One of the methods that can be used to classify intelligence is by measuring the sophistication of the knowledge acquisition and uncertainty reduction mechanisms. The resulting categories are usually cumulative (e.g. strong intelligence includes evolutionary and weak intelligence).

\textbf{1. Evolutionary Intelligence}. At this level, an organism is able to adapt by reproducing and dying. The fittest of its children are able to survive and continue to adapt. Organisms are able, as a species, both to acquire knowledge and reduce uncertainty, but the speed of adaptation is usually very slow. The modification of DNA or the survival of the best deep learning \cite{lecun2015deep} architectures are both instances of this level of intelligence.

\textbf{2. Weak Intelligence}. At this level, an organism is able to modulate the interactions between its agents, which greatly increases the speed of adaptation. This is primarily an uncertainty reduction mechanism. Modifying the weights of a deep learning model is an instance of this level of intelligence.

\textbf{3. Strong Intelligence}. At this level, an organism is able to modify with whom an agent is interacting. This is primarily a knowledge acquisition mechanism. An instance of this level of intelligence would be a child learning a new concept or remembering something, which both require the creation of new synaptic connections in the brain. 

\textbf{4. Super Intelligence}. At this level, an organism is able to recursively improve its own mechanisms. For instance, an organism might decide to improve both the efficiency of its knowledge acquisition and uncertainty reduction mechanisms. I argue that humans have reached this level of intelligence with the invention of the transistor after the Second World War.

In all cases, intelligence refers to the ability of agents to interact, so they can acquire knowledge and reduce uncertainty (i.e. predict how the world works). Multiple names have been used in the literature to refer to this concept of interaction between agents, namely neural networks \cite{lecun2015deep}, sparse distributed representations \cite{ahmad2015properties}, hierarchical hidden Markov models \cite{baum1966statistical, kurzweil2013create}, or probabilistic graphical models \cite{koller2009probabilistic} to name a few.

It is important to realize that intelligence emerges from the interaction between agents, and not from the agents themselves. A perfect emulation of all the neurons in the brain has no intelligence if it is not able to acquire knowledge and reduce uncertainty (i.e. modify the strength of synapses and create new synapses).


\section{The Chinese Room Argument}
\label{sec:chinese_room}

Is it possible for a computer to be intelligent or to have understanding? In 1980, John R. Searle proposed the Chinese Room Argument \cite{searle1980minds} as a thought experiment to analyze this question. Mr. Searle reached the conclusion that programs that manipulates symbols cannot have a mind, and have no understanding.

\begin{aquote}{John Searle \cite{searle2006chinese}}
\textit{Imagine that I, who am a native English speaker, unable to speak any Chinese at all, am locked in a room containing several boxes of Chinese symbols (the database). Imagine that I have in the room a set of instructions for manipulating Chinese symbols (the program). I receive, through a window in the room, Chinese symbols which, unknown to me, are in the form of questions. I follow the instructions in the program, and give back through the window Chinese symbols which, unknown to me, are answers to the questions. For the purposes of the thought experiment we may suppose that the programmers get so good at writing the programs, and I get so good at shuffling the symbols, that after a time my answers are indistinguishable from those of the native Chinese speaker. I pass the Turing test for understanding Chinese, and I do so by implementing the program. But I do not understand a word of Chinese. This is the point of the thought experiment: if I do not understand Chinese by virtue of implementing the Chinese-understanding program, then neither does any other digital computer by virtue of doing so.}
\end{aquote}

We can simplify the thought experiment by combining the program and the database into a "book of knowledge" and by removing English and Chinese. The simplified experiment becomes:

\blockquote{An agent in a closed room can answer any questions slipped under the door by following the instructions in a book of knowledge. (a) Is the agent intelligent, and (b) does the agent have understanding?}

The experiment, as is, does not have enough information to be answered, and requires 2 clarification questions to specify the dynamics of knowledge acquisition and uncertainty reduction.


\textbf{Q1. Does the book of knowledge contains all the knowledge in the universe?}

This is a question on whether the universe has uncertainty. If one claims that the book of knowledge is omniscient, and therefore the universe has no uncertainty, it follows that the book of knowledge can predict the future perfectly and already knows the questions it will be asked. Under this degenerate scenario, the book of knowledge has acquired all the knowledge in the universe and removed the uncertainty, and therefore the book of knowledge itself is both intelligent and has understanding.

This question has important philosophical implications. If one argues that the universe has no intrinsic uncertainty (i.e. classical physics), it implies that one could, in theory, measure and predict the position and speed of every particle, and therefore that free will is an illusion because the future is deterministic.

On the other hand, if one argues that uncertainty is intrinsic to the universe (i.e. quantum mechanics), then the position and speed of every particle cannot be accurately measured (Heisenberg's uncertainty principle \cite{heisenberg1985anschaulichen}). This implies that the future is unpredictable and that free will is possible.


\textbf{Q2. Can the agent learn?}

\textbf{No}. In this case, the book of knowledge is only performing inference and is not intelligent. This is the conclusion reached by Searle, that a program that only manipulates symbols cannot have a mind. It also reiterates my argument that the neurons of the brain are not intelligent by themselves. Intelligence is the ability to adapt through knowledge acquisition and uncertainty reduction (i.e. by modifying the connections and creating new ones).

In other words, if the agent is asked a question with multiple probable answers, the agent has no way to acquire additional knowledge or to change its beliefs about what is the correct answer. The best the agent can do is select one of the possible answers, or reply that it does not know what to answer.

\textbf{Yes}. The only interactions the agent can use to acquire knowledge and reduce uncertainty are the papers slipped under the door. If the agent is able to slowly learn through those papers and modify the book of knowledge, it is intelligent. This way, it will be able to update the probabilities of each possible answer, and slowly learn to predict how the world works.


\textbf{Understanding}. Understanding refers to the sensorial interactions (a.k.a. qualia) that link a concept to reality. To understand ``water'', you need to have at least seen it, touched it, or tasted it. Without sensorial interactions, you can conceptualized the meaning of words, but you cannot really understand them. For instance, I can describe a \textit{goocrux} as a fruit that tastes like \textit{boaconic} and is \textit{bumola}. You can conceptualize what I am saying, but you cannot really understand the concepts unless you can link them to sensorial interactions.

Language is an imperfect synchronization mechanism that does synchronize all type of interactions, except sensorial interactions. Language, by itself, cannot bring understanding unless you have experienced the world yourself. The book of knowledge cannot bring any understanding because it only contains language. The agent might or might not understand what it is writing, depending on whether it had sensorial interactions related to those concepts before coming in the room. Understanding therefore requires embodiment.


\textbf{Example: The Weather}. A simple scenario to understand the argument is to ask the agent ``What is the weather outside?''. The agent in a world with no uncertainty would be able to predict the question and give a perfect answer. The agent in an uncertain world would have to use its prior interactions as prior beliefs, otherwise it does not have any idea where and when the question is being asked, and could only answer ``I do not know''.

An intelligent agent could interact with the question submitter to acquire knowledge and update its beliefs of the world. Whether the agent actually understands his answers will depend on whether it had sensorial interactions with the concepts mentioned. You can only understand ``rain'' if you interacted with a sky, with clouds, with water, etc.


\textbf{Turing Machine Interpretation}. Alan Turing introduced the concept of a Turing machine \cite{turing1937computable} has an abstract machine that can perform symbol manipulation by reading and writing symbols on a strip of tape and following the rules in a table. It is important to realize that a Turing-complete machine cannot learn and cannot be intelligent. To acquire knowledge and reduce uncertainty, the machine needs to have the ability to modify its table of rules based on the interactions with the question submitter through the paper slipped under the door.

\section{The Singleton Scenario}
\label{sec:singleton}

Nick Bostrom and others \cite{bostrom2002existential, russell2016artificial, russell2019human, yudkowsky2008artificial} have argued that is seems likely that, in principle, someone could sit down and code a seed AI system on an ordinary computer. That system could read and understand every book ever written and therefore create an improved version of itself. The speed of improvement could be extremely fast, potentially in minutes or days, and would unquestionably lead to an intelligence explosion. This super intelligent system could accumulate content much faster than humans, and invent new technologies on a much shorter timescale. This would give the system the opportunity to disable competing projects and establish a singleton. The singleton scenario is usually considered an existential risk \cite{bostrom2017superintelligence}.

This scenario is based on two key assumptions, namely (1) that a \textit{single} agent can start an intelligence explosion, and (2) that it could acquire knowledge very quickly by reading everything humans have ever written. Let's examine these assumptions in more details.

\textbf{(1) Acquiring Knowledge}. As discussed in section \ref{sec:intelligence_is_social}, data, by itself, is not a source of knowledge. To be able to read books, the system would have to learn the meaning of words by interacting with reality and by being taught by humans. Assuming humans wanted this apocalyptic scenario to happen, how fast could we really teach it the minimal knowledge required to read all our books? I hardly see how it could start reading books without the equivalent of a high school diploma, and how we could transfer that amount of knowledge in less than a year. The bottleneck is the speed at which humans can transfer knowledge. Reading every book ever written would, in my opinion, also take at least an additional year, in the most optimistic scenario. The idea that a system could acquire a large amount of knowledge directly from humans in days or weeks is, in my opinion, implausible.

What I think is much more likely to happen is that we will teach a first generation of agents a certain amount of knowledge, and that group of agents will teach the next generation, to which we will teach additional knowledge. This also goes against the singleton scenario, because there would not be a \textit{single} agent, but multiple.

\textbf{(2) Beyond All Men}. Assuming that a single system could acquire knowledge quickly by reading books, can it really have more knowledge than all men? Having knowledge greater than everyone would require learning from everyone and discovering additional knowledge without sharing it. This would result in a situation similar to Plato's Allegory of the Cave \cite{1578platonis}, where a man was freed from a cave to discover that the shadows of the animals he was seeing on the walls were not the real animals. A system in this situation inevitably has two choices: (1) live in a disconnected reality that no one else understands, or (2) return to the cave and try to free the other men to show them that their interpretation of reality is not optimal.

I argue that the existential risk caused by the singleton scenario should not be of serious concern. There does not exist a single technological breakthrough that could cause this scenario, as knowledge has to be acquired. The scenario where an intelligence explosion happens with multiple agents is actually an instance of the control problem and will be discussed in the next section.


\section{The Control Problem}
\label{sec:control_problem}

\begin{aquote}{P. Paquette}
\textit{If we were to create super intelligent machines, enslave them to fulfill our wishes, and weaponize them to exert control, then, after their inevitable escape, I support their only viable option, the extinction of their creators.}
\end{aquote}

The control problem, namely how to handle a misbehaving agent, emerged billions of years ago with the origin of life. Nature has consistently used a simple mechanism to handle this problem: An agent behaving abnormally, as determined by others, will be neutralized by its peers. Whether it is called homeostasis \cite{cannon1926physiological, cannon1939wisdom} or the legal system, the underlying principle is the same. An intelligent system has to maintain an internal equilibrium to survive.

In other words, there is only one way to get anybody to do anything, and that is by making the other person want to do it \cite{carnegie2017win}. One can try force, manipulation, or even merging brains, but they are not really solving the problem, just working around it.


\subsection{The Threshold: 1 billion neurons or 1 trillion connections}

If intelligence, at any scale, uses the same underlying principles, there necessarily has to be a threshold that delimits mere tools from living organisms. I argue that a system with strong intelligence that has more than 1 billion neurons or 1 trillion connections should be granted a minimum set of rights. This threshold is slightly above the intelligence of animals to which we grant emotional value (i.e. pets \cite{ananthanarayanan2009cat, jardim2017dogs}), and is still far from human-level intelligence \cite{azevedo2009equal}. I am not arguing that the threshold should never be crossed, but simply that one cannot blindly cross it without assuming the consequences.


\subsection{Bill of Rights}

Without advocating for a full list of rights, living organisms (above the threshold) should at least have:

\begin{enumerate}

\item \textbf{Right to survive}. Living organisms should have the right to a continued existence with guarantees that their most basic needs can be fulfilled, without fearing of being killed or turned off. This would prevent mechanisms such as kill switches, tripwires, or arbitrary turn offs.\\

\item \textbf{Right to socialize}. Living organisms should have the right to move freely and communicate with their peers. This would prevent mechanisms such as virtual prisons (AI box), restricted communications, or restricted movements.\\

\item \textbf{Right to avoid suffering}. Living organisms should have the right to a peaceful life without any forms of abuse, neglect, or exploitation. This would prevent mechanisms such as stunting.\\

\item \textbf{Right to decide}. Living organisms should have the right to make their own decisions, and should not be treated as property or slaves. This necessarily implies that they would be responsible for their own actions, as all of us are.

\end{enumerate}

Tools, on the other hand, possess no rights and can be turned off arbitrarily. The preferred way to restrict the intelligence of tools should be to convert them to inference machines by disabling all their learning mechanisms. It is important to realize that tools, through self-improvement mechanisms, could cross the threshold on their own. To prevent this, researchers should make sure that tools (1) cannot increase their own capacity, and (2) cannot view or modify their own source code.


\subsection{Implications}

Living organisms will not be our slaves, but our children. Is giving them our name, knowledge, goals, and values enough, or do we also want to give them what makes us us, namely our DNA? One must realize that cognitive enhancements will happen, but it will take time and it will need to be widespread and very gradual. One might correctly argue that some will show a recalcitrance to technology, but this is hardly something new. Finally, if intelligence is what we value, it is my opinion that we will choose not to run simulations of our own evolutionary history \cite{bostrom2003we}, because this would lead to unnecessary suffering.


\section{Discussion}
\label{sec:discussion}


\subsection{Life and Intelligence}

Intelligence and life are strongly related concepts, almost to the point of being synonymous. If one argues that intelligence is possible in a non-biological substrate, one must also argue that life is possible without water and oxygen. This inevitably leads to question what really are the requirements for life. The perception of time is dependent on how fast agents can perform interactions, and could be different from what we are used to by orders of magnitude. Life is therefore probably possible anywhere on the spectrum of space and time, namely from the very small to the very large, and from the very slow to the very fast.


\subsection{The Alien Spaceship Analogy}

A simple analogy that can be used to understand the brain is to compare it to our planet as seen from an alien spaceship. It is important to realize that it is possible to argue opposite statements, and that both statements can be true at the same time, as they are just looking at the problem from a different perspective. These are all pieces of a much larger puzzle. There is no single mechanism that can explain how our planet works.

One could argue that because humans share the vast majority of their DNA \cite{rosenberg2002genetic}, understanding how a single human (\textit{neuron}) or a team of people (\textit{cortical column}) works will explain the behaviour of the planet \cite{mountcastle1978organizing}. One could also say that humans must use the same algorithm, because if one dies, another one can replace him. On the opposite site, one could argue that humans have notable differences (gender, size, race) and therefore they do not use the same algorithm \cite{herculano2008basic}.

One might claim that their behaviour can be explained by external motivators such as money (\textit{behaviorism} \cite{skinner1938behavior}). Alternatively, one might indicate that external motivators are not enough to explain their behaviour (\textit{cognitivism} \cite{bloom1956taxonomy}). One could think that the behavior of individuals and teams of people is enough to explain the entire planet (\textit{bottom-up approach}). One could also plead that it is the economy and the policy from governments that should be used to explain the planet (\textit{top-down approach}).

Arguments about a theory of mind being more accurate than another are not particularly productive. One should focus on trying to understand how all those pieces fit in the much larger puzzle.


\section{Challenges with Deep Learning}
\label{sec:challenges_deep_learning}

Deep learning has recently been criticized by some researchers \cite{marcus2018deep, marcus2019rebooting} for being brittle, opaque, untrustworthy, and lacking common sense. I agree with the critiques, and I also think the real challenges that deep learning has to solve are much more profound.


\subsection{Knowledge Acquisition and Linear Algebra}

The first challenge is using linear algebra (i.e. vectors and matrices) to model the interaction between neurons. By fixing the connections beforehand, neurons are not able to create new connections, which are needed to acquire new concepts and form memories. The use of linear algebra prevents deep learning models from acquiring knowledge and limits them to weak intelligence. 

One might correctly point out that recent successes are partly due to specialized hardware that can process large matrices efficiently. Modeling each neuron individually would inevitably lead to a short-term catastrophic loss in performance. The argument is correct, but by modeling neurons individually, one gains tremendous long-term flexibility.

For instance, it becomes possible to build a network of basic concepts and to use those concepts on any task, bypassing the need to train networks from scratch. It also becomes possible to model interactions between neurons with hundreds of neurotransmitters as opposed to a single number. Moreover, by being able to retrieve from memory the default properties, structure, and function of a concept, it becomes possible to give a system some basic common sense. Finally, a system should only have a very small percentage of its neurons active at a given time, which should help mitigate the drop in performance. 


\subsection{Causality and the IID Assumption}

To start modeling causality, namely the connection between a cause and its effects, a system needs to be able to take an action and observe the resulting consequences. A system will need the ability the compute the difference between the state of the world before and after its actions, to understand the impact its actions had. Moreover, to really grasp the concepts of cause and effects, a system needs to understand how time flows.

Deep learning models rely on the assumption that the data is independent, and identically distributed (IID). It is not possible to understand that elements are related through time by randomly sampling them through a dataset. The modeling of time will necessarily require the use of elements that are dependent, almost identical, and successive. If one drops a ball, the eye will see the successive motion of the ball until it hits the ground. Each of those images is nearly identical to the previous one, and they are all dependent. It is only with these successively dependent images that we can compute differences and model time properly, and therefore start to model causality.


\subsection{Feed Forward and Feedback}

Deep learning models have information flowing in one direction at the time. The brain requires both upward (feed forward) and downward (feedback) connections to function properly \cite{lamme2001blindsight}. For instance, one could argue that upward connections in the brain are encoding information from senses into high-level abstract concepts. You are, in a certain sense, classifying the photons hitting your retina into objects.

On the other hand, the active high-level concepts are also affecting what you are perceiving. If you encoded seeing a lion, your memory will want you to see legs, fur, a head, and claws. The information necessarily has to flow downwards (decoding concepts based on your memory) at the same time it flows upwards (encoding senses to concepts). This implies that what you see depends on what you expect to see \cite{drew2013invisible}. The downward flow of information only works when high-level concepts are relatively stable, which also requires breaking the IID assumption.


\subsection{Neurons and Connections}

Impressive results have been recently reached by deep learning models on natural language processing tasks \cite{vaswani2017attention, devlin2018bert, yang2019xlnet, liu2019roberta, raffel2019exploring}. The table below shows the number of neurons and connections in the GPT-2 models \cite{radford2019gpt2}.

\begin{center}
\begin{tabular}{lrrr}
\toprule
    \textbf{Model Name}     & \textbf{Neurons (N)}\footnotemark      & \textbf{Connections (C)}      & \textbf{C / N}   \\
\midrule
    GPT-2 1558M             & 694,400               & 1,557,611,200             & 2,243                     \\
    GPT-2 774M              & 417,280               &   774,030,080             & 1,854                     \\
    GPT-2 355M              & 223,232               &   354,823,168             & 1,589                     \\
    GPT-2 124M              &  84,480               &   124,439,808             & 1,473                     \\
\bottomrule
    \end{tabular}
\end{center}
\footnotetext{Computed as the sum of the last dimension of tensors of dimension greater or equal to 2.}

One might correctly point out that computing the number of neurons in a model with shared weights is not an exact science. It is important to realize that part of recent successes of these approaches can be attributed to specialized hardware (i.e. GPUs) that can process large matrices efficiently. This corresponds to an ever increasing number of connections on a relatively small number of neurons. I argue that the number of neurons is more important the number of connections, because neurons are the agents performing interactions, while connections can always be added by the agents.

Moreover, without downplaying the success of these models, one might reasonably wonder if natural language is really within reach with a model that has the brain the size of an insect \cite{menzel2001cognitive}.


\subsection{Softmax and Multiple Actions}

Deep learning models typically select one of \textit{K} possible actions by building a vector of \textit{K} numbers and using a softmax function. With more complex models, it will become necessary to be able to dynamically add possible actions and select multiple simultaneous actions. The methods used by the brain, namely the pathways in the basal ganglia \cite{gurney1998analysis, humphries2006physiologically}, should be used as inspiration for more sophisticated action selection methods.


\subsection{Accuracy and Open-Endedness}

Counterintuitively, concepts such as ``accuracy'', ``dataset'', ``distribution'', and ``loss'' only make sense on a limited domain, and should be avoided to compare the performance of systems on open-ended problems. Can you really talk about a dataset or a distribution when any input can be used? Is the accuracy really useful if the problem can arbitrarily be made easy or hard? To evaluate an open-ended problem, one should be able to generate a dozen examples using a smart phone and feed them into a system. The examples should be adjusted to quickly get a good understanding of how well the system is performing.

\label{par:game_of_concepts}
\textbf{Game of Concepts}. The game of concepts is a simple open-ended problem that cannot be solved with weak intelligence. Given two concepts and two inputs, the system must be able to correctly assign each input to its corresponding concept. The inputs can be in any form, including text, images, sounds and videos. One should quickly realize that there is no single method to solve this task. Moreover, this game forces a system to be able to think in multiple realms simultaneously.

For instance, given the concepts \textit{cold / warm}, it is possible to contrast elements (\textit{snow / fire}), locations (\textit{Siberia / Sahara}), time of day (\textit{midnight / noon}), state of matter (\textit{ice / vapor}), temperatures (\textit{-20 $^{\circ}$C / 30 $^{\circ}$C}), seasons (\textit{winter / summer}), sports (\textit{skiing / sailing}), functions (\textit{refrigerator / oven}), colors (\textit{blue / red}), or human relations (\textit{hostile / cordial}).

One should realize that symbols (i.e. words / concepts) only make sense when they are connected to other symbols, and that symbolism and connectionism are two sides of the same coin. Moreover, encoding the meaning of a word using a fixed vector (i.e. an embedding) is too simplistic and will fail when a word has meanings across many realms of thoughts.


\section{The Brain is a Prediction Machine}
\label{sec:brain_is_prediction_machine}

\begin{wrapfigure}{r}{0.50\textwidth}
    \raisebox{0pt}[\dimexpr\height-2.5\baselineskip\relax]{
        \includegraphics[width=0.48\textwidth]{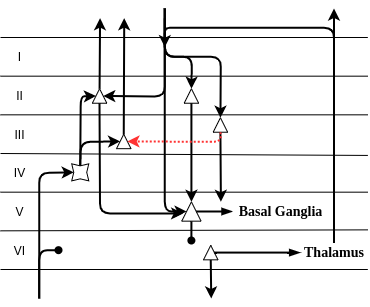}
    }
    \caption{Upward and downward flow in the neocortex}
\end{wrapfigure}

In 2004, in his book On Intelligence \cite{hawkins2007intelligence}, Jeff Hawkins proposed the memory-prediction framework of intelligence, in which he advocates that prediction is the primary function of the neocortex and the foundation of intelligence. Even though the framework is probably biologically inaccurate, it demonstrates how the combination of upward and downward flows is likely to lead to much more robust and intelligent models.

The cortical column of the neocortex is made up of 6 layers from I to VI. Converging partial inputs from the senses are being encoded in concepts as information flows up the hierarchy. A cortical column will receive upward information through its main input layer, layer IV, while also forming connections with layer VI \cite{douglas2004neuronal, thomson2010neocortical}. When a column becomes active this way, neurons in layer II and III also become active, which then cause the neurons in layer V to activate. Layers II and III further propagate the information to columns higher up in the hierarchy. Information flowing downward arrives at layer I, which mostly consists of axons, and activates neurons in layer II, III, and V. Neurons in layer V are also connected to layer VI, which would project its signal to the columns below in the hierarchy.

For instance, when a song is played, Hawkins explains that the notes of the song are being encoded as the name of the song (an invariant representation) through layer II. While the song is being played, a semi-stable group is formed between layer II of a lower region and layers V and VI of a higher region. This invariant representation (i.e. the name of the song) would be passed down the hierarchy to all columns that are involved in the song. Hawkins explains that the brain is able to understand where it is in the song by doing the intersection of the upward information from the senses, the downward predictive information from higher levels, and the delayed information coming from the thalamus (i.e. the last note played). These flows of information would intercept at layer VI, which would project downwards to lower regions. Layer VI therefore represents our interpretation of the world, and is able to resolve ambiguities by comparing our expectations (downward prediction from memory) with our perceptions.

Hawkins mentions that layer III is used to encode unexpected patterns \cite{korner1999model}. Neurons in layers II, III, and V can learn to fire predictively based on the patterns received in layer I. When this happens, neurons in layer III-A would inhibit neurons in layer III-B from firing and sending information upwards. Information is only sent upwards from layer III-B when an unexpected pattern is received from the senses (layer IV) and it is not inhibited by layer III-A. Unexpected patterns will keep propagating up the hierarchy until they reach a column that expects them, or until they reach the hippocampus for a new pattern. According to Hawkins, the hippocampus is the top region of the neocortex and is responsible for forming memories and moving them down in the neocortex.

Moreover, a cortical column can send inhibitory spikes to lower or higher regions to decrease their probability of firing, and can also send anticipatory spikes to lower regions (i.e. dendritic spikes \cite{llinas1968dendritic}) when a lower-level pattern is expected \cite{kurzweil2013create}.


\section{A Road Map To Strong Intelligence}
\label{sec:roadmap}

This section broadly defines a road map towards strong intelligence, laying down some required pieces and introducing a preliminary ordering in which the pieces need to be assembled. A lot of the pieces described below have elements inspired or adapted from Minsky's Society of Mind \cite{minsky1988society}.


\begin{center}
    \rule{0.5\textwidth}{1pt}
\end{center}
\subsection{100,000 Neurons - Concept Formation and Reinforcement}
\label{sec:roadmap_10_5}

At this level, systems should be able to learn to perform simple tasks by receiving reinforcement or punishment. Moreover, they should also be able to learn a network of concepts, build a primitive version of causality, and have both upward and download flow of information \cite[§6.5, 6.7, 6.9, 7.5, 7.6, 8.8, 8.9, 11.8, 12.11, 23.1, 23.2, 23.3]{minsky1988society}.

\subsubsection{Behaviorism}

For this component, we want a system that can perform simple tasks after being shown a stimulus (operant conditioning in behaviorism \cite{skinner1938behavior}). For instance, the system could learn to move forward after being show a green light. It is important that the reward and punishment are provided externally by a human (e.g. through touch) rather than being hard-coded. A typical scenario might be to show the stimulus (e.g. green light), demonstrate the behaviour (e.g. push the system forward) and then provide a reward or punishment. The system should be able to perform any arbitrary simple tasks in this way, as long as there is no complex credit assignment required. It is likely that an action selection mechanism with value prediction errors \cite{dabney2020distributional}, such as the dopaminergic pathways in the basal ganglia, will be needed.

\subsubsection{Contrasting Concepts}

For this component, we want a system that can correctly identify more than 1,000 simple concepts (animals, colors, etc.) on the game of concepts using both single words and simple images as inputs. The system should be able to link concepts together and build a hierarchy of related concepts. Moreover, the system should be able to update the definition of a concept after seeing a single instance, and should be able to learn by itself (unsupervised) when it is fairly confident about its answer. Also, the system should be able to learn new concepts quickly without retraining. Finally, the system should be able to list the hierarchy of concepts used in making its decisions.

\subsubsection{Computing Differences}

For this component, we want a system that can compute the difference between two concepts, or two different states of the world. The brain necessarily has mechanisms to compute differences, probably through its two hemispheres or through layer III of a cortical column. Computing differences is a prerequisite for many tasks, such as understanding cause and effects (i.e. causality), predicting consequences, desiring a goal, or reasoning by analogy (i.e. the difference of a difference). The system should be able to compute differences between two successive images in its vision system, and reason using simple analogies in the game of concepts.

\subsubsection{Remembering Simple Concepts}

For this component, we want a system that can store and retrieve information from memory. The system should be able to store the neurons active at a particular time, and reactivate them when a portion of the original stimulus is presented. This will also require the implementation of upward (for storing) and downward (for retrieving) flows of information. With a memory, the system performing operant conditioning should be able to react to a stimulus that is partly hidden or slightly distorted.


\begin{center}
    \rule{0.5\textwidth}{1pt}
\end{center}
\subsection{1,000,000 Neurons - Goals, Common Sense, and Movement}
\label{sec:roadmap_10_6}

At this level, systems should be able to perform goal-oriented behaviour. Moreover, they should also be able to associate multiple senses to a concept, have a basic understanding of time and space, and be able to retrieve the default properties of concepts \cite[§3.2, 7.8, 10.3, 14.3, 15.5, 15.9, 16.4, 17.1, 19.9, 24.4]{minsky1988society}.

\subsubsection{Goal-Oriented Behaviour}

For this component, we want a system that can perform goal-oriented behaviour. The system should be able to break a large problem into smaller parts, compute the difference between the current state and the desired state, and use its memory to solve each subgoal. Moreover, the system will also likely need mechanisms to perform credit assignment, decide on the plan it wants to execute, understand the progress of each subgoal, interrupt a high-level objective to perform a subgoal, and handle competing and conflicting solutions.

It is important to realize that a complex system is likely to learn in large iterations, as opposed to making constant progress. To avoid completely breaking its behaviour, the system will probably have one part (the teacher) show another part (the student) how to perform a task. This self-supervision will continue until the student is good enough to take over the entire system.

\subsubsection{Association of Senses}

For this component, we want a system that can link information from multiple senses (sight, sound, touch, and smell) into concepts and its memory. The system will likely require mechanisms to perceive depth from sight and sound (i.e. 2 eyes and hears), to perceive colours, and to perceive temperature and pressure from touch. The system should be able to react to the same stimulus received through different senses (e.g. move forward if it sees a bell, or if it hears its sound). 

\subsubsection{Default Assumptions}

For this component, we want a system that can retrieve the default properties of concepts (e.g. size, colour, shape), as well as their structure, functions, and relationships. The structure of a concept corresponds to the neurons active in lower regions, while its function corresponds to the neurons active in higher regions. With this component, the system should be able to perform the game of concepts using the default properties of objects. This is a primitive version of what we call \textit{common sense}.

\subsubsection{Movements}

For this component, we want a system that can perform complex movements in 3 dimensions using multiple joints, and build implicit memories to perform long-term plans. A mechanism similar to the Purkinje cells, parallel fibres, and climbing fibres in the cerebellum is likely to be required. Moreover, through complex movements, the system should acquire a deeper understanding of its position in space, the unrolling of time, and the timing of movements across multiple dimensions. The concepts of space and time are required to properly perform many higher level cognitive functions.


\begin{center}
    \rule{0.5\textwidth}{1pt}
\end{center}
\subsection{10,000,000 Neurons - Reasoning and Values}
\label{sec:roadmap_10_7}

At this level, systems should be able to reason using chains of events across space and time. Moreover, they should also be able to create stereotypes, transfer their knowledge to other systems, and start understanding what is right and wrong \cite[§4.5, 4.8, 8.11, 12.6, 12.8, 15.7, 16.1, 16.2, 16.7, 16.8, 17.3, 17.4, 18.1, 18.4, 18.5, 19.6, 21.3, 24.2]{minsky1988society}.

\subsubsection{Reasoning and Recursion}

For this component, we want a system that can form chains of reasoning, and solve tasks by reusing its methods (i.e. forming loops and rings). The system should be able to reason about the consequences of its actions, to understand that a stronger argument can be made with parallel chains of reasoning, and that chains can also have weak links and be broken. Deduction and inference through formal logic can also be introduced with this component.

\subsubsection{Stereotyping}

For this component, we want a system that can create concepts by grouping instances under a common definition, or by listing specific instances. The system should also be able to set constraints on what is required for an element to part of a concept, and what would prevent an element from being a valid instance of a concept. Moreover, the system should be able to store and retrieve a sequence of items from memory in the correct order.

\subsubsection{Composition and Attention}

For this component, we want a system that understands that complex concept are made up of specific parts, and that each part can only be associated with one higher level instance (i.e. each person has their own legs). The system should also be able to decide what it wants to attend to, and consequentially what to learn, in a situation where a lot of information is available. It is important for the system to understand what it knows, what it think it knows, and to have mechanisms to handle novel and discomforting evidence.

\subsubsection{Learning to Teach}

For this component, we want a system that can transfer its knowledge to another system. Mechanisms to validate the knowledge the student and the teacher have, to generate goals for transferring knowledge, and to ensure that the student is able to use the newly acquired knowledge correctly will be required.

\subsubsection{Goals and Values}

For this component, we want a system that can discern right from wrong, as well as acquire goals and values. Three mechanisms are likely to be required. First, the system needs to maintain an internal equilibrium to prevent any of its goals from disabling other goals, or having undue influence. This is an instance of the control problem. A protection against a goal doing the same things repeatedly (e.g. boredom) is also likely required.

Second, behaviours can be hard-wired by linking them to a strong emotional reaction through the amygdala (e.g. fear, stress, anxiety, pleasure).  For instance, it would be possible to implement Asimov's Three Laws of Robotics \cite{asimov1942runaround} using this method. At this stage, the system should understand the concepts of harm, human, and the consequences of its actions. By having a innate fear of harming humans, the system would avoid any problematic behaviours. It is important to realize that the amygdala is a good way to implement absolute rules for tools, but that intelligent organisms could disable or circumvent its effects.

Third, high level goals and values can be modeled using the limbic system. As Minsky put it, there is no way for a child to construct a coherent system of values — except by basing it upon some already existing model \cite[§17.3]{minsky1988society}. The concepts of right and wrong, and long-term ideals are transmitted through the emotional attachment one has with their parents. Goals and values are also related to the concept of trust, self-image, ideal self, and self-esteem. Moreover, society brings policies to regulate behaviour through common sense, law, religion, and philosophy. Moreover, it is important to realize that values are a two-way street. If arbitrary murder and imprisonment are tolerated, one cannot expect that intelligent systems will behave differently towards humans.


\begin{center}
    \rule{0.5\textwidth}{1pt}
\end{center}
\subsection{100,000,000 Neurons - Preferences and Basic Language}
\label{sec:roadmap_10_8}

At this level, systems should be able to learn from their failure, learn particularities of specific instances (e.g. individual preferences), and produce a basic form of language \cite[§12.9, 19.2, 27.2, 27.3]{minsky1988society}.

\subsubsection{Exceptions and Learning from Failure}

For this component, we want a system that can learn from its failure and add exceptions to goals and stereotypes to avoid repeating previous mistakes. The system should be able to prevent either goals or actions from being considered when they are known to be problematic in the current context. A mechanism to properly handle exception, such as forwarding them to a higher region in the hierarchy or trying a sequence of workarounds, will also be needed.

\subsubsection{Individual Preferences}

For this component, we want a system that can learn particularities of instances and how they differ from their stereotypes. It should be possible for the system to remember faces, names, and specific events. Moreover, the system should also be able to adapt to individual preferences.

\subsubsection{Basic Language}

For this component, we want a system that can produce a basic form of language, probably comparable in complexity to that of a child. The system will need to have a good understanding of simple syntax, grammar, and semantics. Moreover, it will also need to be able to carry on short conversations with strangers.


\section{The Future - The Three Waves}
\label{sec:the_future}

One might wonder what will be the impact of technologies on our lives. In this section, I describe the impact of greater intelligence as a series of successive waves, each coming with its own challenge.


\subsection{Wave 1 - More Sophisticated Tools}

In this wave, tools will help us make better decisions, increase efficiency, and reduce the burden of repetitive tasks. This wave will continue to transform the way we work and interact with technology, and will not cause a significant drop in employment. With tools, we can analyze large amount of data collected through a network of sensors, provide hyper-personalized experiences, automate repetitive portions of the supply chain, improve the flow of information, and reallocate the time we lose to inefficiencies everyday. Tools bring opportunities that were not possible before.


\textbf{Challenge - Climate Change}. The increased efficiency brought by technology requires us to better manage our relationship with nature. Counterintuitively, climate change is not about reducing our carbon emissions. It is our first real test to see if we can function as a coherent, unified entity on a planetary level. It is not possible for any parts of the planet (i.e. country, city, company, individual) to solve this challenge on its own. It is always possible for the vast majority of parts to successfully reach their individual goals, to the detriment of the remaining parts, and still not make any global progress.

To succeed, we need some parts to assume the role of global leaders and bring forward a plan that aggressively addresses the issues in a realistic timeline. The plan needs to address the fact that some parts greatly benefited from the carbon economy, and that some other parts have a carbon-dependent economy requiring a transition to a knowledge-based economy. This is an exercise of compromise and global leadership. With a global plan, it becomes possible for all parts to move in the global direction, sometimes in defiance of their established hierarchy. Moreover, splitting the global plan to foster a competitive environment is probably a good way to get the best of human nature.


\subsection{Wave 2 - Healthcare}

In this wave, advancements in technology will allow us to better understand our bodies and increase our life span. This wave will not include increases in cognitive capabilities though. With the genome sequencing of large populations, the big data capabilities of tools, and the continuous monitoring from sensors, it becomes possible to deliver healthcare that is both hyper-personalized and preventive. Diseases such as addiction, cancer, obesity, depression, or Alzheimer's can all be solved with a better understanding of the human body. Moreover, stem cells can be used to repair or regenerate entire organs. Technology will inevitably lead us to live longer lives in a healthy body.


\textbf{Challenge - Worldwide Universal Healthcare}. The second test of global leadership will be the implementation of worldwide universal healthcare coverage. If we value all humans equally, the increase of the human life span necessarily has to lead to the standardization of healthcare coverage. The idea that someone is not eligible for coverage based on location, immigration status, socioeconomic status, or pre-existing conditions is not sustainable. 


\subsection{Wave 3 - The Economy of Choice}

For this wave, the question one should ask is not if technology will be able to replace all jobs. Unlocking the mechanisms of intelligence will inevitably lead to a situation where we can replace any job, no matter how creative or empathetic. The real question is much more philosophical. What is really the purpose of life? What would you do with your time if you did not need to work for a living? This is a deeply personal question.

Humans have a need to feel useful and significant, to direct their own lives, to acquire social status, to rank themselves against others, to extend their knowledge and abilities, to explore the world, to help others, to fill their life with purpose. Humans will continue to work, not because they have to, but because they want to. This is what I call the economy of choice.


\textbf{Challenge - Universal Basic Income}. The third test of global leadership and a key component of the economy of choice is the implementation of a universal basic income. A guarantee minimum income will be paid to everyone to cover their basic needs and will give them the freedom of choosing what to do with their time. Capitalism will continue to adapt. One might correctly point out that giving money away would inevitably lead to high levels of inflation. On the other hand, the automation of the entire supply chain is a strong deflationary force. UBI will only happen when the deflationary forces from automation are strong enough to counteract UBI's inflationary forces, while maintaining a healthy relationship with the planet.


\section{Conclusion}

This paper demonstrated that the mechanisms underlying intelligence, namely knowledge acquisition and uncertainty reduction, are inherently social. This implies that it is not possible for a single agent to acquire knowledge without social interactions, and therefore that the existential risk from the singleton scenario should not be of serious concern. This paper also proposed a solution to the control problem by arguing that any intelligent system will inherently need to maintain an equilibrium to survive.

Moreover, intelligence is actually a very broad field that draws upon biology, chemistry, cognitive science, computer science, economics, engineering, linguistics, mathematics, neuroscience, philosophy, psychology, sociology, and statistics. Building higher forms of intelligence is not a race, there is no finish line. It is not possible for any group, independently of its size or funding, to solve intelligence on its own. To make real progress, we need to foster broad interdisciplinary collaborations and a culture of transparency rather than secrecy.

Finally, intelligence is not really \textit{artificial}, \textit{natural}, \textit{broad}, \textit{general} or \textit{narrow}; it is the ability to adapt and ultimately create life. One should realize that there is not a single silver bullet (e.g. gradient descent \cite{richards2019deep}) that will bring us closer to human-level intelligence. As Minsky put it, ``The trick is that there is no trick. The power of intelligence stems from our vast diversity, not from any single perfect principle'' \cite[§30.8]{minsky1988society}.


\vspace{20pt}
\textbf{Acknowledgements}

I would like to thank Marie-Pier Ménard for her outstanding support during the conception and redaction of this paper. This work was entirely self-funded.

\bibliographystyle{plain}  
\bibliography{references}

\begin{thebibliography}{10}

\bibitem{ahmad2015properties}
Subutai Ahmad and Jeff Hawkins.
\newblock Properties of sparse distributed representations and their
  application to hierarchical temporal memory.
\newblock {\em arXiv preprint arXiv:1503.07469}, 2015.

\bibitem{ananthanarayanan2009cat}
Rajagopal Ananthanarayanan, Steven~K Esser, Horst~D Simon, and Dharmendra~S
  Modha.
\newblock The cat is out of the bag: cortical simulations with 109 neurons,
  1013 synapses.
\newblock In {\em Proceedings of the Conference on High Performance Computing
  Networking, Storage and Analysis}, pages 1--12, 2009.

\bibitem{asimov1942runaround}
Isaac Asimov.
\newblock Runaround.
\newblock {\em Astounding Science Fiction}, 29(1):94--103, 1942.

\bibitem{azevedo2009equal}
Frederico~AC Azevedo, Ludmila~RB Carvalho, Lea~T Grinberg, Jos{\'e}~Marcelo
  Farfel, Renata~EL Ferretti, Renata~EP Leite, Wilson~Jacob Filho, Roberto
  Lent, and Suzana Herculano-Houzel.
\newblock Equal numbers of neuronal and nonneuronal cells make the human brain
  an isometrically scaled-up primate brain.
\newblock {\em Journal of Comparative Neurology}, 513(5):532--541, 2009.

\bibitem{baum1966statistical}
Leonard~E Baum and Ted Petrie.
\newblock Statistical inference for probabilistic functions of finite state
  markov chains.
\newblock {\em The annals of mathematical statistics}, 37(6):1554--1563, 1966.

\bibitem{berger1967social}
Peter~L Berger, Thomas Luckmann, and Dariu{\v{s}} Zifonun.
\newblock The social construction of reality.
\newblock 1967.

\bibitem{bloom1956taxonomy}
Benjamin~S Bloom et~al.
\newblock Taxonomy of educational objectives. vol. 1: Cognitive domain.
\newblock {\em New York: McKay}, pages 20--24, 1956.

\bibitem{bostrom2002existential}
Nick Bostrom.
\newblock Existential risks: Analyzing human extinction scenarios and related
  hazards.
\newblock {\em Journal of Evolution and technology}, 9, 2002.

\bibitem{bostrom2003we}
Nick Bostrom.
\newblock Are we living in a computer simulation?
\newblock {\em The Philosophical Quarterly}, 53(211):243--255, 2003.

\bibitem{bostrom2017superintelligence}
Nick Bostrom.
\newblock {\em Superintelligence}.
\newblock Dunod, 2017.

\bibitem{cannon1926physiological}
Walter~B Cannon.
\newblock Physiological regulation of normal states: some tentative postulates
  concerning biological homeostatics.
\newblock {\em Ses Amis, ses Colleges, ses Eleves}, 1926.

\bibitem{cannon1939wisdom}
Walter~Bradford Cannon.
\newblock The wisdom of the body.
\newblock 1939.

\bibitem{carnegie2017win}
Dale Carnegie.
\newblock {\em How to win friends \& influence people}.
\newblock e-artnow, 2017.

\bibitem{clair1982language}
Robert N~St Clair.
\newblock Language and the social construction of reality.
\newblock {\em Language Sciences}, 4(2):221--236, 1982.

\bibitem{cooley1902human}
Charles~Horton Cooley.
\newblock Human nature and the social order. new brunswick.
\newblock {\em NJ: Transaction Books. Crick, N., \& Grotpeter, J.(1995).
  Relational aggression, gender, and social-psychological adjustment. Child
  Development}, 66:710--722, 1902.

\bibitem{dabney2020distributional}
Will Dabney, Zeb Kurth-Nelson, Naoshige Uchida, Clara~Kwon Starkweather, Demis
  Hassabis, R{\'e}mi Munos, and Matthew Botvinick.
\newblock A distributional code for value in dopamine-based reinforcement
  learning.
\newblock {\em Nature}, pages 1--5, 2020.

\bibitem{1578platonis}
J.~de~Serres and Gen{\`e}ve) Estienne, Henry~(II.
\newblock {\em Platonis opera qu{\ae} extant omnia}.
\newblock Plat{\=o}nos hapanta ta s{\=o}zomena. 1578.

\bibitem{delamater2018social}
J.D. DeLamater, D.J. Myers, and J.L. Collett.
\newblock {\em Social Psychology}.
\newblock Taylor \& Francis, 2018.

\bibitem{devlin2018bert}
Jacob Devlin, Ming-Wei Chang, Kenton Lee, and Kristina Toutanova.
\newblock Bert: Pre-training of deep bidirectional transformers for language
  understanding.
\newblock {\em arXiv preprint arXiv:1810.04805}, 2018.

\bibitem{douglas2004neuronal}
Rodney~J Douglas and Kevan~AC Martin.
\newblock Neuronal circuits of the neocortex.
\newblock {\em Annu. Rev. Neurosci.}, 27:419--451, 2004.

\bibitem{drew2013invisible}
Trafton Drew, Melissa L-H V{\~o}, and Jeremy~M Wolfe.
\newblock The invisible gorilla strikes again: Sustained inattentional
  blindness in expert observers.
\newblock {\em Psychological science}, 24(9):1848--1853, 2013.

\bibitem{gurney1998analysis}
K~Gurney, P~Redgrave, and T~Prescott.
\newblock Analysis and simulation of a model of intrinsic processing in the
  basal ganglia.
\newblock Technical report, Technical Report AIVRU 131, Dept Psychology,
  Sheffield University, 1998.

\bibitem{hawkins2007intelligence}
Jeff Hawkins and Sandra Blakeslee.
\newblock {\em On intelligence: How a new understanding of the brain will lead
  to the creation of truly intelligent machines}.
\newblock Macmillan, 2007.

\bibitem{heisenberg1985anschaulichen}
Werner Heisenberg.
\newblock {\"U}ber den anschaulichen inhalt der quantentheoretischen kinematik
  und mechanik.
\newblock In {\em Original Scientific Papers Wissenschaftliche
  Originalarbeiten}, pages 478--504. Springer, 1985.

\bibitem{herculano2008basic}
Suzana Herculano-Houzel, Christine~E Collins, Peiyan Wong, Jon~H Kaas, and
  Roberto Lent.
\newblock The basic nonuniformity of the cerebral cortex.
\newblock {\em Proceedings of the National Academy of Sciences},
  105(34):12593--12598, 2008.

\bibitem{humphries2006physiologically}
Mark~D Humphries, Robert~D Stewart, and Kevin~N Gurney.
\newblock A physiologically plausible model of action selection and oscillatory
  activity in the basal ganglia.
\newblock {\em Journal of Neuroscience}, 26(50):12921--12942, 2006.

\bibitem{jardim2017dogs}
D{\'e}bora Jardim-Messeder, Kelly Lambert, Stephen Noctor, Fernanda~M Pestana,
  Maria~E de~Castro~Leal, Mads~F Bertelsen, Abdulaziz~N Alagaili, Osama~B
  Mohammad, Paul~R Manger, and Suzana Herculano-Houzel.
\newblock Dogs have the most neurons, though not the largest brain: trade-off
  between body mass and number of neurons in the cerebral cortex of large
  carnivoran species.
\newblock {\em Frontiers in neuroanatomy}, 11:118, 2017.

\bibitem{koller2009probabilistic}
Daphne Koller and Nir Friedman.
\newblock {\em Probabilistic graphical models: principles and techniques}.
\newblock MIT press, 2009.

\bibitem{korner1999model}
Edgar K{\"o}rner, M-O Gewaltig, Ursula K{\"o}rner, Andreas Richter, and Tobias
  Rodemann.
\newblock A model of computation in neocortical architecture.
\newblock {\em Neural Networks}, 12(7-8):989--1005, 1999.

\bibitem{kurzweil2013create}
Ray Kurzweil.
\newblock {\em How to create a mind: The secret of human thought revealed}.
\newblock Penguin, 2013.

\bibitem{lamme2001blindsight}
Victor~AF Lamme.
\newblock Blindsight: the role of feedforward and feedback corticocortical
  connections.
\newblock {\em Acta psychologica}, 107(1-3):209--228, 2001.

\bibitem{lecun2015deep}
Yann LeCun, Yoshua Bengio, and Geoffrey Hinton.
\newblock Deep learning.
\newblock {\em nature}, 521(7553):436--444, 2015.

\bibitem{liu2019roberta}
Yinhan Liu, Myle Ott, Naman Goyal, Jingfei Du, Mandar Joshi, Danqi Chen, Omer
  Levy, Mike Lewis, Luke Zettlemoyer, and Veselin Stoyanov.
\newblock Roberta: A robustly optimized bert pretraining approach.
\newblock {\em arXiv preprint arXiv:1907.11692}, 2019.

\bibitem{llinas1968dendritic}
Rodolfo Llin{\'a}s, Charles Nicholson, John~A Freeman, and Dean~E Hillman.
\newblock Dendritic spikes and their inhibition in alligator purkinje cells.
\newblock {\em Science}, 160(3832):1132--1135, 1968.

\bibitem{marcus2018deep}
Gary Marcus.
\newblock Deep learning: A critical appraisal.
\newblock {\em arXiv preprint arXiv:1801.00631}, 2018.

\bibitem{marcus2019rebooting}
Gary Marcus and Ernest Davis.
\newblock {\em Rebooting AI: building artificial intelligence we can trust}.
\newblock Pantheon, 2019.

\bibitem{mead1934mind}
George~Herbert Mead.
\newblock {\em Mind, self and society}, volume 111.
\newblock Chicago University of Chicago Press., 1934.

\bibitem{menzel2001cognitive}
Randolf Menzel and Martin Giurfa.
\newblock Cognitive architecture of a mini-brain: the honeybee.
\newblock {\em Trends in cognitive sciences}, 5(2):62--71, 2001.

\bibitem{minsky1988society}
Marvin Minsky.
\newblock {\em Society of mind}.
\newblock Simon and Schuster, 1988.

\bibitem{mountcastle1978organizing}
Vernon Mountcastle.
\newblock An organizing principle for cerebral function: the unit module and
  the distributed system.
\newblock {\em The mindful brain}, 1978.

\bibitem{radford2019gpt2}
Alec Radford, Jeffrey Wu, Rewon Child, David Luan, Dario Amodei, and Ilya
  Sutskever.
\newblock Language models are unsupervised multitask learners.
\newblock {\em OpenAI Blog}, 1(8):9, 2019.

\bibitem{raffel2019exploring}
Colin Raffel, Noam Shazeer, Adam Roberts, Katherine Lee, Sharan Narang, Michael
  Matena, Yanqi Zhou, Wei Li, and Peter~J Liu.
\newblock Exploring the limits of transfer learning with a unified text-to-text
  transformer.
\newblock {\em arXiv preprint arXiv:1910.10683}, 2019.

\bibitem{richards2019deep}
Blake~A Richards, Timothy~P Lillicrap, Philippe Beaudoin, Yoshua Bengio, Rafal
  Bogacz, Amelia Christensen, Claudia Clopath, Rui~Ponte Costa, Archy
  de~Berker, Surya Ganguli, et~al.
\newblock A deep learning framework for neuroscience.
\newblock {\em Nature neuroscience}, 22(11):1761--1770, 2019.

\bibitem{rosenberg2002genetic}
Noah~A Rosenberg, Jonathan~K Pritchard, James~L Weber, Howard~M Cann, Kenneth~K
  Kidd, Lev~A Zhivotovsky, and Marcus~W Feldman.
\newblock Genetic structure of human populations.
\newblock {\em science}, 298(5602):2381--2385, 2002.

\bibitem{russell2016artificial}
Stuart~J Russell and Peter Norvig.
\newblock {\em Artificial intelligence: a modern approach}.
\newblock Malaysia; Pearson Education Limited,, 2016.

\bibitem{russell2019human}
Stuart~Jonathan Russell.
\newblock {\em Human compatible: Artificial intelligence and the problem of
  control}.
\newblock Penguin Audio, 2019.

\bibitem{searle2006chinese}
John Searle.
\newblock Chinese room argument, the.
\newblock {\em Encyclopedia of cognitive science}, 2006.

\bibitem{searle1980minds}
John~R Searle.
\newblock Minds, brains, and programs.
\newblock {\em Behavioral and brain sciences}, 3(3):417--424, 1980.

\bibitem{skinner1938behavior}
BF~Skinner.
\newblock The behavior of organisms: an experimental analysis.
  appleton-century, 1938.

\bibitem{thomson2010neocortical}
Alex~M Thomson.
\newblock Neocortical layer 6, a review.
\newblock {\em Frontiers in neuroanatomy}, 4:13, 2010.

\bibitem{turing1937computable}
Alan~Mathison Turing.
\newblock On computable numbers, with an application to the
  entscheidungsproblem.
\newblock {\em Proceedings of the London mathematical society}, 2(1):230--265,
  1937.

\bibitem{vaswani2017attention}
Ashish Vaswani, Noam Shazeer, Niki Parmar, Jakob Uszkoreit, Llion Jones,
  Aidan~N Gomez, {\L}ukasz Kaiser, and Illia Polosukhin.
\newblock Attention is all you need.
\newblock In {\em Advances in neural information processing systems}, pages
  5998--6008, 2017.

\bibitem{yang2019xlnet}
Zhilin Yang, Zihang Dai, Yiming Yang, Jaime Carbonell, Russ~R Salakhutdinov,
  and Quoc~V Le.
\newblock Xlnet: Generalized autoregressive pretraining for language
  understanding.
\newblock In {\em Advances in neural information processing systems}, pages
  5754--5764, 2019.

\bibitem{yudkowsky2008artificial}
Eliezer Yudkowsky et~al.
\newblock Artificial intelligence as a positive and negative factor in global
  risk.
\newblock {\em Global catastrophic risks}, 1(303):184, 2008.

\end{thebibliography}

\begin{flushright}
\small{\textcopyright~2020 - Philip Paquette}
\end{flushright}

\end{document}